\documentclass[11pt,a4paper]{article}
\usepackage[utf8]{inputenc}
\usepackage[T1]{fontenc}
\usepackage{lmodern}
\usepackage[margin=1in]{geometry}
\usepackage{booktabs}
\usepackage{hyperref}
\usepackage{enumitem}
\usepackage{sectsty}
\usepackage{parskip}
\usepackage{amsmath}
\usepackage{amssymb}
\usepackage{tikz}
\usepackage[super,numbers,sort&compress]{natbib}
\usetikzlibrary{positioning, arrows.meta}

\title{Grounding Investor Views: Neural Predicates in the Black-Litterman Model}
\author{Marcos Florencio \\
        Rio de Janeiro, Brazil \\
        \texttt{marcos.florencio@puc-rio.br}}
\date{\today}
\begin{document}
\maketitle

\begin{abstract}
Portfolio construction under the Black-Litterman model requires investors to specify views on asset returns alongside explicit uncertainty estimates---a process that remains largely subjective and difficult to scale. We propose a formal approach in which neural predicates serve as a structured, probabilistic mechanism for view generation. In our formulation, structured financial analysis data is processed through a compositional hierarchy of neural predicates whose outputs---probability distributions over market stances---are mapped to the pick matrix $\mathbf{P}$, the view return vector $\mathbf{q}$, and the view uncertainty matrix $\boldsymbol{\Omega}$ of the Black-Litterman model. View confidence is derived from predicate output distributions, providing a data-driven alternative to subjective uncertainty elicitation. The resulting approach is interpretable, in the sense that any portfolio weight can be traced back through the predicate's logical chain to the underlying data, and fully differentiable, enabling end-to-end learning.
\end{abstract}

\vspace{1em}

\section{Introduction}
\label{sec:introduction}

Portfolio construction involves decision-making under uncertainty. The mean-variance approach of \citet{markowitz1952} established the importance of portfolio covariance, but its sensitivity to expected return estimates often produces unstable and implausible allocations. The Black-Litterman model \citep{black1991, black1992} addresses this issue by combining equilibrium returns implied by the CAPM with investor views held at explicit confidence levels, yielding more stable and diversified portfolios.\footnote{The CAPM \citep{sharpe1964, lintner1965} is used for exposition. The approach developed here is agnostic to this choice.}

However, Black-Litterman specifies how to incorporate views without explaining how they should be generated. In practice, directions, magnitudes, and confidence levels are typically assigned through subjective judgment, limiting reproducibility and scalability \citep{idzorek2004}. Rich analytical information from valuation models, earnings analysis, and balance-sheet diagnostics is often reduced to ad hoc point estimates with arbitrary uncertainty assumptions.

This paper proposes a theoretical approach in which \emph{neural predicates} 
\citep{manhaeve2018} generate Black-Litterman views. Their output 
distributions over discrete stances (e.g., bullish, neutral, bearish) 
naturally provide view direction, magnitude, and uncertainty. We make three 
contributions: (i) a formal mapping from neural predicate outputs to the 
Black-Litterman components $\mathbf{P}$, $\mathbf{q}$, and 
$\boldsymbol{\Omega}$; (ii) an entropy-based measure of uncertainty that 
replaces fixed covariance heuristics; and (iii) a compositional method for 
combining multiple predicates into a coherent asset view.

\section{Background and Related Work}
\label{sec:related}

\subsection{Portfolio Optimization and the Limits of Mean-Variance}

Modern portfolio theory originates with \citet{markowitz1952}, who cast portfolio selection as a quadratic program: given expected returns $\boldsymbol{\mu}$ and covariance $\boldsymbol{\Sigma}$, the investor seeks weights $\mathbf{w}$ that minimize variance for a given expected return, tracing the efficient frontier.

The approach suffers a well-documented fragility: optimal weights are extraordinarily sensitive to the inputs, particularly to $\boldsymbol{\mu}$ \citep{michaud1989}. Because expected returns must be estimated from noisy data, estimation error propagates directly into weights. \citet{jobson1982} showed that in-sample optimality rarely survives out of sample, and \citet{best1991} demonstrated analytically that small perturbations in return estimates produce large, economically implausible swings in weights. The optimizer thus acts as an \emph{error maximizer}, concentrating weight in assets whose returns are most overestimated \citep{michaud1989}.

The literature has responded with shrinkage estimators \citep{ledoit2004}, resampling \citep{michaud1989}, and robust optimization \citep{goldfarb2003}. These improve stability but share a limitation: they address estimation error statistically, without incorporating the domain knowledge needed to form return expectations---a requirement left to the implementation. The Black-Litterman model instead addresses the source of the fragility, replacing unconstrained estimation with a structured Bayesian approach anchored to a stable prior.

\subsection{Incorporating Qualitative Judgment in Quantitative Models}

Research on incorporating qualitative judgment into portfolio construction generally follows three approaches: direct use of analyst forecasts, factor models, and Bayesian methods such as Black-Litterman. Analyst recommendations and price targets contain predictive information \citep{barber2008}, but they provide point estimates without a principled measure of uncertainty and are difficult to aggregate systematically. Factor models \citep{fama1993} offer a structured approach based on systematic risk premia, yet they are less suited to firm-specific insights and rarely propagate estimation uncertainty into portfolio decisions. The Black-Litterman model \citep{black1991, black1992} addresses these issues by combining equilibrium returns with investor views and explicitly modeling view uncertainty; extensions have incorporated non-normal returns \citep{meucci2006} and non-linear views \citep{meucci2010}. Despite their differences, all three approaches rely on an informal translation of qualitative judgment into quantitative inputs, sacrificing reproducibility and the structure of the underlying analysis. As financial data become increasingly rich and high-dimensional, a systematic mechanism for generating Black-Litterman-compatible views is needed. We argue that neural predicates provide such a mechanism.

\subsection{Neuro-Symbolic AI and the Rise of Neural Predicates}

AI has historically been divided between symbolic approaches, which emphasize rule-based reasoning and interpretability \citep{newell1976}, and connectionist approaches, which learn distributed representations from data \citep{rumelhart1986}. Their complementary strengths and weaknesses---precision and compositionality on one side, robustness and scalability on the other---motivated the development of neuro-symbolic systems \citep{fodor1988}. Early work showed that neural computation could implement logical operations \citep{mcculloch1943}, while later systems such as KBANN \citep{towell1994}, CILP \citep{garcez1999}, and CILP++ \citep{franca2014} demonstrated how symbolic knowledge could guide neural architectures and be recovered from trained models. Neural predicates, introduced in DeepProbLog by \citet{manhaeve2018}, provide a prominent solution: logical predicates whose truth probabilities are generated by neural networks through neural annotated disjunctions. This integration enables probabilistic inference, symbolic reasoning, and end-to-end learning within a unified approach.

\subsection{Prior Work on Machine Learning in Portfolio Construction}

Research on machine learning in portfolio construction broadly spans return prediction, alternative data, and tighter integration with optimization. Studies such as \citet{gu2020} and \citet{chen2019} show that neural networks can outperform traditional models in forecasting cross-sectional returns, but they generally provide point estimates without a principled treatment of uncertainty. A parallel literature uses NLP to extract signals from news, analyst reports, earnings calls, and social media \citep{tetlock2007, loughran2011, huang2018, chen2014}, with recent extensions employing large language models \citep{lopez-lira2025}; however, converting these signals into Black-Litterman views remains largely ad hoc. Other work combines machine learning with portfolio optimization through robust optimization, reinforcement learning, and deep hedging \citep{ban2018, zhang2020, carbonneau2020}, while extensions of Black-Litterman incorporate quantitative signals into the view vector $\mathbf{q}$ \citep{he1999}. Yet no existing approach provides a formal and interpretable mapping from structured, multi-dimensional analysis to the full $(\mathbf{P}, \mathbf{q}, \boldsymbol{\Omega})$ representation, and the specification of $\boldsymbol{\Omega}$ in particular remains weakly grounded, often reduced to a fixed multiple of the prior covariance \citep{idzorek2004}.

\section{Neural Predicates and the Black-Litterman Model}
\label{sec:predicates}

\subsection{Symbolic Reasoning and Neural Predicates}

Symbolic AI models intelligent behavior as rule-governed manipulation of discrete, interpretable structures, its dominant formalism being \emph{first-order logic} (FOL) \citep{russell2010}. A \emph{predicate} $P$ of arity $n$ is a function $P : \mathcal{D}^n \to \{\top, \bot\}$ over a domain $\mathcal{D}$; an \emph{atom} such as $\mathit{bullish}(x)$ is the smallest expression to which a truth value can be assigned, and a variable-free atom obtained by substituting constants (\emph{grounding}) is a \emph{ground atom}---e.g. $\mathit{bullish}(\mathit{apple})$, asking whether Apple is in a bullish stance. Truth-preserving \emph{inference rules}, chief among them modus ponens (from $\varphi \rightarrow \psi$ and $\varphi$, derive $\psi$), then derive new facts, each conclusion carrying an explicit justification chain. Systems built this way succeeded in well-structured domains---theorem proving, expert systems, Prolog \citep{kowalski1974}---but cannot handle perceptual data, noise, or learning.

The obstacle is representational. A neural network computes a continuous function $f_{\boldsymbol{\theta}} : \mathbb{R}^d \to \mathbb{R}^k$ over distributed, non-interpretable activations, whereas symbolic systems operate on discrete structures in which a rule either fires or does not. This creates the \emph{symbol grounding problem} \citep{harnad1990}: a symbolic system cannot determine the truth of $\mathit{undervalued}(x)$ from raw data, so grounding must be stipulated externally by a human---the principal bottleneck as data grow. Compounding it, discrete logical operations are step functions with zero gradient almost everywhere, so embedding symbolic reasoning in a training loop breaks backpropagation and precludes joint end-to-end learning. Real decision-making---including, we argue, generating investor views---requires both perception (where neural networks excel) and compositional reasoning (where symbolic systems excel).

A neural predicate \citep{manhaeve2018} is a logical predicate whose extension is determined by a neural network rather than by enumeration. Let $q$ have possible output values $\mathbf{u} = (u_1, \ldots, u_m)$ over input terms $\mathbf{t}$. A \emph{neural annotated disjunction} (nAD) is

\begin{equation}
    \mathtt{nn}(m_q,\, \mathbf{t},\, u,\, \mathbf{u}) 
    \mathrel{::} 
    q(\mathbf{t}, \, u),
    \label{eq:nad}
\end{equation}

with semantics that the probability of the ground atom $q(\mathbf{t}, u_i)$ equals the $i$-th network output,

\begin{equation}
    P\bigl(q(\mathbf{t}, u_i) = \top\bigr)
    = \bigl[f_{m_q}(\mathbf{t})\bigr]_i,
    \quad 
    \sum_{i=1}^{m} \bigl[f_{m_q}(\mathbf{t})\bigr]_i = 1,
    \label{eq:nad_semantics}
\end{equation}

where $f_{m_q} : \mathcal{X} \to \Delta^{m-1}$ maps inputs to the probability simplex via a softmax layer. For financial stance classification, $\mathtt{nn}(f_{\mathit{stance}}, x, s, \{\mathit{bullish}, \mathit{bearish}, \mathit{neutral}\}) :: \mathit{stance}(x, s)$, with $x$ encoding a company's profile. DeepProbLog \citep{manhaeve2018} embeds nADs into ProbLog \citep{deraedt2007}, computing query probabilities by \emph{weighted model counting} (WMC); because network outputs appear as weights and WMC can be made differentiable, gradients back-propagate into the network, enabling end-to-end learning.

\subsection{Compositional Reasoning}
\label{subsec:compositional}

The power of neural predicates lies in composing them through logical rules, which distinguishes such a system from a collection of independent classifiers. Atomic predicates form the perceptual layer---for instance $\mathit{valuation}(x,v)$, $\mathit{earnings}(x,e)$, and $\mathit{health}(x,h)$, each an nAD over its own categories---and rules compose them:

\begin{align}
    \mathit{stance}(x, \mathit{bullish}) 
        &\leftarrow
        \mathit{valuation}(x, \mathit{under}),\,
        \mathit{earnings}(x, \mathit{strong}),\,
        \mathit{health}(x, \mathit{healthy}), \label{eq:bullish_comp}\\
    \mathit{stance}(x, \mathit{bearish}) 
        &\leftarrow
        \mathit{valuation}(x, \mathit{over}),\,
        \mathit{earnings}(x, \mathit{weak}). 
        \label{eq:bearish_comp}
\end{align}

The derived $\mathit{stance}(x, s)$ is not a network but a logical consequence, its probability computed by the inference engine. Treating each network output as the probability of the corresponding ground atom, under ProbLog's independence assumption rule \eqref{eq:bullish_comp} fires with probability

\begin{equation}
    P(\mathit{bullish}\text{ via }\eqref{eq:bullish_comp})
    = P(\mathit{val.}{=}\mathit{under})\cdot
      P(\mathit{earn.}{=}\mathit{strong})\cdot
      P(\mathit{health}{=}\mathit{healthy}),
    \label{eq:rule_prob}
\end{equation}

and the total probability of each stance sums such terms over deriving rules, with corrections for overlap \citep{deraedt2007}, yielding the output distribution

\begin{equation}
    \bigl(
        P(\mathit{bullish}\mid x),\;
        P(\mathit{bearish}\mid x),\;
        P(\mathit{neutral}\mid x)
    \bigr).
    \label{eq:stance_dist}
\end{equation}

This architecture is \emph{interpretable by construction}: each probability in \eqref{eq:stance_dist} traces back through explicit logical steps to the atomic predicate outputs, so the engine can report which rules contributed and with what weight. Unlike post-hoc attention or gradient attribution, the explanation is not a surrogate but the actual computational path---consequential where investment committees require justifiable reasoning and regulation may mandate explainability.

\subsection{Neural Predicates as View Generators}

The central conceptual claim is that the output \eqref{eq:stance_dist}, denoted $\boldsymbol{\pi}_i \in \Delta^2$ for company $i$, is structurally isomorphic to a Black-Litterman view triplet $(\mathbf{p}_k, q_k, \Omega_k)$. The dominant stance determines the direction ($\mathbf{p}_k$): bullish a positive absolute view, bearish a negative one, neutral no view. A mapping from stance probabilities to excess returns, developed in Section~\ref{sec:approach}, yields the view return ($q_k$). And the Shannon entropy of the distribution,

\begin{equation}
    H(\boldsymbol{\pi}_i) =
    -\sum_{s} P(s \mid x_i) \log P(s \mid x_i),
    \label{eq:entropy}
\end{equation}

measures its uncertainty---low for a concentrated (confident) distribution, high for a diffuse one---and maps directly to the view variance ($\Omega_k$), a data-driven alternative to the ad hoc conventions detailed below. The correspondence reflects a deeper alignment: both approaches represent beliefs as distributions, update them against evidence, and propagate uncertainty explicitly rather than collapsing it to a point. The former generates the beliefs the latter incorporates.

\subsection{The Black-Litterman Model}
\label{sec:bl}

The Black-Litterman model \citep{black1991, black1992} combines the information in market equilibrium with subjective views to form expected return estimates. Its central innovation is the prior. Rather than estimating returns directly---which amplifies estimation error and destabilizes weights (Section~\ref{sec:related})---it begins from market equilibrium. Under the CAPM \citep{sharpe1964, lintner1965}, the market portfolio with weights $\mathbf{w}_{\mathit{mkt}}$ and return covariance $\boldsymbol{\Sigma}$ is mean-variance efficient, so the \emph{implied equilibrium excess returns} follow by reverse optimization,

\begin{equation}
    \boldsymbol{\Pi} = \delta \boldsymbol{\Sigma}
    \mathbf{w}_{\mathit{mkt}},
    \label{eq:equilibrium}
\end{equation}

with market risk-aversion $\delta = (\mathbb{E}[r_m] - r_f)/\sigma_m^2$. The model treats $\boldsymbol{\Pi}$ as the prior mean over expected returns $\boldsymbol{\mu}$:

\begin{equation}
    \boldsymbol{\mu} \sim \mathcal{N}\bigl(
        \boldsymbol{\Pi},\, \tau \boldsymbol{\Sigma}
    \bigr),
    \label{eq:prior}
\end{equation}

where $\tau > 0$ scales the prior's uncertainty (typically $\tau \in [0.01, 0.05]$; \citealp{black1992}). This prior is grounded in market prices rather than finite-sample noise and, absent views, recommends the well-diversified market portfolio---a default historical mean-variance optimization almost never produces \citep{he1999}. Departures are driven entirely by views.

\subsection{Views and the Posterior}
\label{subsec:bl_posterior}

The model supports $K$ views, encoded in the pick matrix $\mathbf{P} \in \mathbb{R}^{K \times N}$, the view return vector $\mathbf{q} \in \mathbb{R}^K$, and the diagonal uncertainty matrix $\boldsymbol{\Omega} \in \mathbb{R}^{K \times K}$. Row $k$ of $\mathbf{P}$ specifies the portfolio to which view $k$ applies: an \emph{absolute} view has a single entry of $1$; a \emph{relative} view has positive entries summing to $1$ and negative entries summing to $-1$. Each $q_k$ gives the expected return on that portfolio, and each $\omega_k > 0$ its variance (small for high confidence). The views are noisy observations of the prior,

\begin{equation}
    \mathbf{q} = \mathbf{P}\boldsymbol{\mu} + \boldsymbol{\varepsilon},
    \qquad
    \boldsymbol{\varepsilon} \sim \mathcal{N}(\mathbf{0}, \boldsymbol{\Omega}),
    \label{eq:view_model}
\end{equation}

with uncorrelated errors. Bayes' theorem then yields a Gaussian posterior with closed-form mean \citep{he1999}

\begin{equation}
    \boldsymbol{\mu}_{\mathit{BL}} = \boldsymbol{\Pi} +
    \tau\boldsymbol{\Sigma}\mathbf{P}^\top
    \bigl(\mathbf{P}\tau\boldsymbol{\Sigma}\mathbf{P}^\top +
    \boldsymbol{\Omega}\bigr)^{-1}
    \bigl(\mathbf{q} - \mathbf{P}\boldsymbol{\Pi}\bigr),
    \label{eq:bl_posterior_mean_alt}
\end{equation}

the equilibrium return plus a correction proportional to the \emph{view surprise} $\mathbf{q} - \mathbf{P}\boldsymbol{\Pi}$; assets covered by no view retain their equilibrium returns. The scalar $\tau$ sets the prior's weight against the views and is difficult to calibrate \citep{idzorek2004}; we treat it as given. With posterior covariance $\mathbf{M} = [(\tau\boldsymbol{\Sigma})^{-1} + \mathbf{P}^\top\boldsymbol{\Omega}^{-1}\mathbf{P}]^{-1}$, the total return covariance $\boldsymbol{\Sigma} + \mathbf{M}$ feeds a mean-variance optimization

\begin{equation}
    \mathbf{w}^* = 
    \operatorname*{arg\,max}_{\mathbf{w}}
    \Bigl\{
        \mathbf{w}^\top \boldsymbol{\mu}_{\mathit{BL}} -
        \tfrac{\delta}{2}\, \mathbf{w}^\top
        (\boldsymbol{\Sigma} + \mathbf{M})\mathbf{w}
    \Bigr\},
    \label{eq:bl_mvo}
\end{equation}

with unconstrained solution

\begin{equation}
    \mathbf{w}^* = 
    \tfrac{1}{\delta}
    (\boldsymbol{\Sigma} + \mathbf{M})^{-1}
    \boldsymbol{\mu}_{\mathit{BL}}.
    \label{eq:bl_weights}
\end{equation}

Absent views, $\boldsymbol{\mu}_{\mathit{BL}} = \boldsymbol{\Pi}$ gives $\mathbf{w}^* = \mathbf{w}_{\mathit{mkt}}$, so every deviation is attributable to a specific view; and because $\boldsymbol{\mu}_{\mathit{BL}}$ is a precision-weighted average, the equilibrium prior acts as a regularizer, mitigating the error-amplification pathology of Section~\ref{sec:related}.

\subsection{The View Specification Problem}
\label{subsec:view_spec_problem}

The model incorporates views rigorously but offers no method for \emph{generating} them or \emph{quantifying} their uncertainty---its principal practical limitation. Views originate from informal judgment \citep{black1992, he1999}: the multi-dimensional content of an analysis collapses into a scalar $q_k$ by an undocumented, irreproducible process on which two analysts may disagree. The uncertainty $\boldsymbol{\Omega}$ is more acute still, demanding calibrated beliefs about the reliability of one's own judgment. The most widely used approach \citep{idzorek2004} sets

\begin{equation}
    \boldsymbol{\Omega} =
    \alpha \cdot \mathbf{P}(\tau \boldsymbol{\Sigma}) \mathbf{P}^\top,
    \label{eq:idzorek_omega}
\end{equation}

with an ad hoc scalar $\alpha$ that reflects only the analyst's willingness to assert a number and imposes the generally unjustified assumption that view uncertainty is proportional to return volatility. Manual elicitation of a triplet $(\mathbf{p}_k, q_k, \omega_k)$ per asset also does not scale, and systematic strategies address scale only by sacrificing view structure (Section~\ref{sec:related}). The sophistication of the posterior \eqref{eq:bl_posterior_mean_alt} thus rests on inputs of uncertain quality---precisely what Section~\ref{sec:approach} addresses, deriving directions from the dominant stance, magnitudes from a principled stance-to-return mapping, and uncertainties from the output distribution's entropy \eqref{eq:entropy}, each explicit, reproducible, and grounded in the analysis.

\section{Incorporating Neural Predicates into Black-Litterman}
\label{sec:approach}

\subsection{Overview of the Proposed Approach}

\begin{figure}[ht]
\centering
\resizebox{\textwidth}{!}{%
\begin{tikzpicture}[
    node distance=1.5cm and 1.0cm,
    box/.style={rectangle, draw, rounded corners,
                minimum width=2.2cm, minimum height=1.0cm,
                align=center, font=\footnotesize},
    arr/.style={-Stealth, thick}
]
\node[box] (data)   {Company\\Analytical Data\\$\mathbf{x}_i$};
\node[box, right=of data]  (pred)
    {Neural\\Predicate System\\$\boldsymbol{\pi}_i$};
\node[box, right=of pred]  (view)
    {View Matrices\\$\mathbf{P},\,\mathbf{q},\,\boldsymbol{\Omega}$};
\node[box, right=of view]  (bl)
    {BL Posterior\\$\boldsymbol{\mu}_{\mathit{BL}}$};
\node[box, right=of bl]    (port)
    {Portfolio\\Weights\\$\mathbf{w}^*$};

\draw[arr] (data) -- (pred);
\draw[arr] (pred) -- (view);
\draw[arr] (view) -- (bl);
\draw[arr] (bl)   -- (port);
\end{tikzpicture}%
}
\caption{End-to-end architecture of the proposed approach.
Structured analytical data for each asset is processed by a
compositional neural predicate system to produce a probability
distribution over market stances. This distribution is mapped
to the Black-Litterman view triplet
$(\mathbf{P}, \mathbf{q}, \boldsymbol{\Omega})$, which is
combined with the equilibrium prior via the Bayesian update
of Section~\ref{subsec:bl_posterior} to produce the posterior
expected return vector $\boldsymbol{\mu}_{\mathit{BL}}$,
from which optimal portfolio weights are derived.}
\label{fig:architecture}
\end{figure}
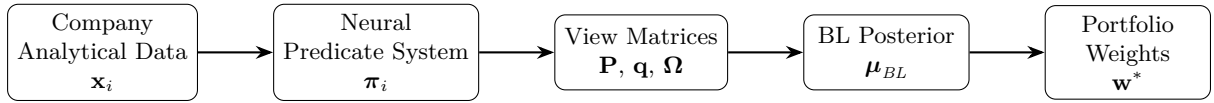

The approach interposes a neural predicate system between raw company-level data and the Black-Litterman view interface, proceeding through four stages (Figure~\ref{fig:architecture}).

\paragraph{Stage 1: Analytical data.}
For each asset $i$, a structured profile $\mathbf{x}_i \in \mathcal{X}$ encodes the multi-dimensional outputs of financial analysis---valuation, earnings quality, balance-sheet diagnostics, and any other relevant structured quantities. Crucially, $\mathbf{x}_i$ retains the full structure of the analysis rather than collapsing it to a scalar; translating that structure into a probabilistic judgment is the predicate system's task.

\paragraph{Stage 2: Neural predicate evaluation.}
The profile is passed through a compositional neural predicate system (Section~\ref{sec:predicates}), which evaluates a hierarchy of atomic predicates and composes their outputs into a stance distribution

\begin{equation}
    \boldsymbol{\pi}_i =
    \bigl(P(\mathit{bullish} \mid \mathbf{x}_i),\;
    P(\mathit{bearish} \mid \mathbf{x}_i),\;
    P(\mathit{neutral} \mid \mathbf{x}_i)\bigr)
    \in \Delta^2.
    \label{eq:predicate_output_fw}
\end{equation}

This vector is the sufficient statistic for view generation: all of $\mathbf{x}_i$ relevant to the update is captured in these three dimensions.

\paragraph{Stage 3: View matrix construction.}
The distribution $\boldsymbol{\pi}_i$ is mapped to the view triplet, developed in Subsections~\ref{subsec:q_mapping}--\ref{subsec:aggregation}, yielding $K \leq N$ views encoded in $\mathbf{P} \in \mathbb{R}^{K \times N}$, $\mathbf{q} \in \mathbb{R}^K$, and $\boldsymbol{\Omega} \in \mathbb{R}^{K \times K}$.

\paragraph{Stage 4: Bayesian update and portfolio construction.}
The triplet enters the Black-Litterman update \eqref{eq:bl_posterior_mean_alt} to produce $\boldsymbol{\mu}_{\mathit{BL}}$, then the mean-variance optimization \eqref{eq:bl_mvo} to produce weights $\mathbf{w}^*$. This stage is entirely standard; the approach's contribution lies in Stage 3.

\subsection{Mapping Neural Predicate Outputs to View Returns}
\label{subsec:q_mapping}

The view return $q_i$ is a scalar expected excess return, which we now derive from $\boldsymbol{\pi}_i$.

\paragraph{Stance-return correspondence.}
We introduce three stance return parameters,

\begin{equation}
    r_{\mathit{bullish}} > 0, \quad
    r_{\mathit{bearish}} < 0, \quad
    r_{\mathit{neutral}} \approx 0,
    \label{eq:stance_returns}
\end{equation}

encoding the investment hypothesis: a bullish assessment implies positive excess return $r_{\mathit{bullish}}$, a bearish one negative return $|r_{\mathit{bearish}}|$, and a neutral one near zero. The natural choice for $r_{\mathit{neutral}}$ is the equilibrium return $\Pi_i$, reflecting that a neutral assessment adds nothing beyond what the market has priced; a predicate assigning equal probability to all stances then gravitates toward the prior, and the posterior return sits near $\Pi_i$.

\paragraph{The view return mapping.}
We define the view return as the expected stance return under the predicate distribution:

\begin{equation}
    q_i = P(\mathit{bullish} \mid \mathbf{x}_i)\cdot
    r_{\mathit{bull}} +
    P(\mathit{bearish} \mid \mathbf{x}_i)\cdot
    r_{\mathit{bear}} +
    P(\mathit{neutral} \mid \mathbf{x}_i)\cdot
    \Pi_i.
    \label{eq:q_mapping}
\end{equation}

This mapping is linear in the predicate probabilities, so $q_i$ varies smoothly as the distribution shifts; it recovers $r_{\mathit{bullish}}$, $r_{\mathit{bearish}}$, or $\Pi_i$ under certainty; and a uniform distribution yields $q_i = (r_{\mathit{bull}} + r_{\mathit{bear}} + \Pi_i)/3$, a moderate return the uncertainty mechanism of Subsection~\ref{subsec:omega_mapping} will downweight.

\paragraph{Calibration of stance return parameters.}
The parameters $r_{\mathit{bullish}}$ and $r_{\mathit{bearish}}$ represent the average excess return of a correct bullish or bearish assessment and can be estimated by regressing realized excess returns on past stance probabilities. Their estimation is left to the implementation; here we treat them as given.

\paragraph{Thresholding and view inclusion.}
Not every output should generate a view: a near-uniform distribution carries little directional information and would add overhead and potential numerical instability. We introduce a \emph{view inclusion threshold} $\gamma \in (0, \frac{1}{3})$, including a view for asset $i$ only if

\begin{equation}
    \max_s P(s \mid \mathbf{x}_i) \geq \frac{1}{3} + \gamma,
    \label{eq:threshold}
\end{equation}

i.e. only if some stance is materially above the uniform baseline. Larger $\gamma$ yields fewer, higher-confidence views; smaller $\gamma$ yields more views at the cost of noisier signals.

\subsection{Constructing the Pick Matrix from Predicate Structure}

The compositional structure of the predicate system provides a natural basis for $\mathbf{P}$, and both absolute and relative views arise from predicate outputs.

\paragraph{Absolute views from single-asset predicates.}
When a predicate applied to asset $i$ alone satisfies the threshold \eqref{eq:threshold}, it generates an \emph{absolute view} asserting return $q_i$ for that asset, with

\begin{equation}
    [\mathbf{P}]_{ki} = 1, \quad
    [\mathbf{P}]_{kj} = 0 \text{ for all } j \neq i.
    \label{eq:absolute_view}
\end{equation}

This is the approach's most direct use: each company's profile generates one absolute view.

\paragraph{Relative views from comparative predicates.}
Given assets $i$ and $j$ with outputs $\boldsymbol{\pi}_i$ and $\boldsymbol{\pi}_j$, a \emph{relative view} asserts the differential

\begin{equation}
    q_{ij} = q_i - q_j,
    \label{eq:relative_view_return}
\end{equation}

with pick vector $[\mathbf{P}]_{k,i} = 1$, $[\mathbf{P}]_{k,j} = -1$, others zero---useful in long-short or market-neutral strategies. More generally, given a bullish set $\mathcal{L}$ and bearish set $\mathcal{S}$, a group-relative view sets $[\mathbf{P}]_{k,i} = 1/|\mathcal{L}|$ for $i \in \mathcal{L}$ and $[\mathbf{P}]_{k,j} = -1/|\mathcal{S}|$ for $j \in \mathcal{S}$, expressing that the average bullish asset outperforms the average bearish one by $q_{\mathcal{L}} - q_{\mathcal{S}}$.

\paragraph{Predicate structure and pick matrix sparsity.}
A naive implementation generates one view per asset, making $\mathbf{P}$ the identity---a valid special case. But the compositional structure permits richer, semantically meaningful multi-asset views: a macro predicate firing on a whole sector can generate a single view on the equal-weighted sector portfolio, leaving within-sector allocation to asset-level predicates. This hierarchical structure---macro predicates for broad views, micro predicates for idiosyncratic ones---mirrors the multi-level reasoning institutional investors employ.

\subsection{Deriving View Uncertainty from Predicate Confidence}
\label{subsec:omega_mapping}

The uncertainty matrix $\boldsymbol{\Omega}$ is the component existing practice handles least satisfactorily; as established in Subsection~\ref{subsec:view_spec_problem}, setting it proportional to the prior covariance \citep{idzorek2004} has no connection to analytical confidence. This subsection is the paper's primary theoretical contribution: a principled, data-driven mapping from the dispersion of the predicate output to the view uncertainty $\omega_i$.

\paragraph{Entropy as a measure of predicate uncertainty.}
The Shannon entropy of $\boldsymbol{\pi}_i$ is

\begin{equation}
    H(\boldsymbol{\pi}_i) = 
    -\sum_{s \in \mathcal{S}}
    P(s \mid \mathbf{x}_i) \log P(s \mid \mathbf{x}_i),
    \label{eq:entropy_fw}
\end{equation}

with $\mathcal{S} = \{\mathit{bullish}, \mathit{bearish}, \mathit{neutral}\}$ and natural logarithms. It is zero at maximum certainty (probability one on a single stance) and maximal at $\log 3 \approx 1.099$ under the uniform distribution, making it a natural measure of the predicate's epistemic confidence.

\paragraph{The uncertainty mapping.}
We propose

\begin{equation}
    \omega_i = \omega_{\min} + (\omega_{\max} - \omega_{\min})
    \cdot \frac{H(\boldsymbol{\pi}_i)}{\log 3},
    \label{eq:omega_mapping}
\end{equation}

with bounds $0 < \omega_{\min} < \omega_{\max}$. The normalized entropy $H(\boldsymbol{\pi}_i)/\log 3 \in [0,1]$ interpolates linearly: a concentrated predicate ($H = 0$) gives $\omega_i = \omega_{\min}$, a diffuse one ($H = \log 3$) gives $\omega_{\max}$. Since $\omega_i$ is monotone increasing in entropy, the posterior weights confident assessments more and uncertain ones less---exactly the Bayesian principle that evidence quality governs evidence weight.

\paragraph{Relation to the prior covariance.}
To keep $\omega_i$ commensurate with the prior covariance $\tau\boldsymbol{\Sigma}$, we calibrate the bounds against its diagonal:

\begin{equation}
    \omega_{\min} = \alpha_{\min} \cdot
    \tau [\boldsymbol{\Sigma}]_{ii}, \qquad
    \omega_{\max} = \alpha_{\max} \cdot
    \tau [\boldsymbol{\Sigma}]_{ii},
    \label{eq:omega_bounds}
\end{equation}

with $0 < \alpha_{\min} < \alpha_{\max}$ dimensionless. This preserves the scale-invariance of the update---more volatile assets get larger view uncertainties, consistent with the intuition that precise views are harder to form on volatile assets. Substituting into \eqref{eq:omega_mapping},

\begin{equation}
    \omega_i = \tau [\boldsymbol{\Sigma}]_{ii}
    \left[\alpha_{\min} + (\alpha_{\max} - \alpha_{\min})
    \cdot \frac{H(\boldsymbol{\pi}_i)}{\log 3}\right],
    \label{eq:omega_final}
\end{equation}

which nests the Idzorek convention \eqref{eq:idzorek_omega} as the special case $H(\boldsymbol{\pi}_i) = 0$ with $\alpha_{\min} = \alpha$. The mapping thus generalizes existing practice by making the confidence parameter a function of the predicate output rather than an analyst-specified constant.

\paragraph{Alternative dispersion measures.}
Other measures may suit specific contexts. The \emph{margin} of the dominant stance,

\begin{equation}
    m_i = \max_s P(s \mid \mathbf{x}_i)
    - \frac{1}{3},
    \label{eq:margin}
\end{equation}

may be more interpretable, while the \emph{Gini impurity} $1 - \sum_s P(s \mid \mathbf{x}_i)^2$ is computationally simpler. Each induces a different functional form for $\omega_i$ but shares the property of mapping concentrated distributions to low uncertainty and diffuse ones to high; the choice is an empirical question left to the implementation.

\subsection{Multi-Predicate Composition and View Aggregation}
\label{subsec:aggregation}

A richer architecture deploys multiple specialized predicates and aggregates their outputs into one view per asset.

\paragraph{Multiple predicate outputs.}
Suppose $L$ predicates, indexed by $\ell$, are applied to asset $i$, each producing $\boldsymbol{\pi}_i^{(\ell)} \in \Delta^2$ and addressing a distinct dimension (valuation, financial health, and so on), with individual view returns and uncertainties

\begin{equation}
    q_i^{(\ell)} = P^{(\ell)}(\mathit{bullish} \mid \mathbf{x}_i)
    r_{\mathit{bullish}} +
    P^{(\ell)}(\mathit{bearish} \mid \mathbf{x}_i) r_{\mathit{bearish}}
    + P^{(\ell)}(\mathit{neutral} \mid \mathbf{x}_i) \Pi_i,
    \label{eq:q_per_pred}
\end{equation}

\begin{equation}
    \omega_i^{(\ell)} = \tau[\boldsymbol{\Sigma}]_{ii}
    \left[\alpha_{\min} + (\alpha_{\max} - \alpha_{\min})
    \cdot \frac{H(\boldsymbol{\pi}_i^{(\ell)})}{\log 3}
    \right].
    \label{eq:omega_per_pred}
\end{equation}

The problem is to combine $\{(q_i^{(\ell)}, \omega_i^{(\ell)})\}_{\ell=1}^L$ into a single $(q_i, \omega_i)$.

\paragraph{Precision-weighted aggregation.}
The natural probabilistic rule weights each view return by its inverse uncertainty:

\begin{equation}
    q_i = \frac{\sum_{\ell=1}^L
    (\omega_i^{(\ell)})^{-1} q_i^{(\ell)}}
    {\sum_{\ell=1}^L (\omega_i^{(\ell)})^{-1}},
    \label{eq:precision_agg}
\end{equation}

with aggregate uncertainty, under independent errors,

\begin{equation}
    \omega_i = \left(\sum_{\ell=1}^L
    (\omega_i^{(\ell)})^{-1}\right)^{-1}.
    \label{eq:precision_omega_agg}
\end{equation}

Equation \eqref{eq:precision_omega_agg} is the standard combination of independent Gaussian estimates: concordant predicates yield a more confident aggregate than any single one. If predicates disagree in direction, their contributions partially cancel in \eqref{eq:precision_agg}, pulling $q_i$ toward neutrality---but \eqref{eq:precision_omega_agg} does not increase under disagreement, since it aggregates precisions regardless of conflict. We address this next.

\paragraph{Disagreement-adjusted uncertainty.}
Directional disagreement is itself informative about assessment difficulty and should raise uncertainty. We augment \eqref{eq:precision_omega_agg} with a penalty:

\begin{equation}
    \omega_i^{\mathit{adj}} = \omega_i \cdot
    \left(1 + \lambda \cdot \mathbb{V}\!\left[
    q_i^{(\ell)}\right]\right),
    \label{eq:disagreement_penalty}
\end{equation}

where $\mathbb{V}[q_i^{(\ell)}] = \frac{1}{L}\sum_\ell (q_i^{(\ell)} - q_i)^2$ is the empirical variance of the individual view returns and $\lambda \geq 0$ scales the penalty. Under agreement ($\mathbb{V} = 0$) the adjusted uncertainty equals \eqref{eq:precision_omega_agg}; under strong disagreement it inflates, letting the update discount the view accordingly.

\paragraph{Hierarchical predicate composition.}
Alternatively, multi-predicate composition can be folded into the logical structure itself (Section~\ref{subsec:compositional}): atomic predicates compose through rules into a derived stance predicate whose distribution $\boldsymbol{\pi}_i$, computed by WMC, feeds \eqref{eq:q_mapping} and \eqref{eq:omega_final} directly, without a separate aggregation step. This has the advantage that the composition rules are explicit and interpretable, but the limitation that they must be specified in advance and may miss relevant interactions. The numerical aggregation of \eqref{eq:precision_agg}--\eqref{eq:disagreement_penalty} is more flexible but less transparent; in practice a hybrid---logical composition for well-understood relationships, numerical aggregation elsewhere---may work best.

\subsection{Theoretical Properties of the Proposed Approach}
\label{subsec:properties}

Consistency, calibration, interpretability, and modularity follow directly from the approach's structure. Consistency arises because the same predicates, rules, and stance parameters \eqref{eq:stance_returns} apply to every asset, ensuring that view returns and uncertainties are comparable across the universe. Calibration is achieved by deriving $\omega_i$ from entropy \eqref{eq:entropy_fw}: uncertain predicates generate larger $\omega_i$ and therefore weaker updates, while confident predicates produce smaller $\omega_i$, a property that follows from the mapping \eqref{eq:omega_final} itself. Interpretability is inherent rather than post hoc, since every portfolio weight can be traced from $\mathbf{x}_i$ through atomic predicate outputs, inference rules, $(q_i,\omega_i)$, and ultimately to $\boldsymbol{\mu}_{\mathit{BL}}$ and the optimal allocation, unlike approximation-based explanations such as saliency methods \citep{lundberg2017}. Finally, the approach is modular: predicate systems, Black-Litterman parameters, and stance-return calibrations can each be updated independently, enabling maintainability, validation, and deployment without retraining the entire system.

\section{Numerical Example}
\label{sec:example}

To illustrate the approach of Section~\ref{sec:approach}, we consider a toy implementation on a two-asset universe using a DeepProbLog inference engine, a large language model as the neural predicate component, and a standard Black-Litterman optimizer. The example is purely demonstrative: the firms are hypothetical, the analytical profiles are stipulated, and no claims about predictive performance are made. Its purpose is simply to show that the proposed mapping is computationally feasible and yields economically interpretable outcomes. The universe contains two companies, \textsc{Acme} and \textsc{Globex}, with parameters reported in Table~\ref{tab:asset_params}. The return correlation is fixed at $\rho = 0.25$, the market risk-aversion coefficient at $\delta = 2.5$, and the stance returns at $r_{\mathit{bull}} = 0.20$, $r_{\mathit{neut}} = 0.00$, and $r_{\mathit{bear}} = -0.15$. These values are chosen for expository simplicity; their empirical calibration is left to the implementation.

\begin{table}[ht]
    \centering
    \caption{Asset-level parameters for the illustrative example.
        Market capitalization weights are used to derive the
        equilibrium implied returns $\boldsymbol{\Pi}$ via
        equation~\eqref{eq:equilibrium}.}
    \label{tab:asset_params}
    \begin{tabular}{lcc}
        \toprule
            Parameter & \textsc{Acme} & \textsc{Globex} \\
        \midrule
            Annualized return volatility $\sigma_i$ & $0.28$ & $0.18$ \\
            Market capitalization (USD) & $5.0 \times 10^9$ & $1.2 \times 10^{10}$ \\
            Market capitalization weight $w_i^{\mathit{mkt}}$ & $0.294$ & $0.706$ \\
        \bottomrule
    \end{tabular}
\end{table}

\subsection*{Neural Predicate Evaluation}

Each company's analytical profile is passed to a neural
predicate implemented as a call to GPT-4o
\citep{openai2024} via the OpenAI API, structured according
to the LangChain wrapper detailed in the public repository. The model is prompted with the
company's structured financial profile and constrained, via
Pydantic schema validation and structured output enforcement,
to return a probability distribution over the three stance
categories summing to exactly 100. Temperature is set to
zero to ensure deterministic outputs, consistent with the
predicate consistency requirement discussed in
Section~\ref{sec:predicates}.

The DeepProbLog inference engine, initialized with the
\texttt{ExactEngine} for exact probabilistic inference,
compiles the logical program and evaluates six ground
queries — one per stance per company — in $6.50$ seconds
of wall-clock time, of which $4.27$ seconds are attributable
to the \textsc{Acme} LLM call and $2.21$ seconds to the
\textsc{Globex} LLM call. The resulting stance distributions
are reported in Table~\ref{tab:predicate_outputs}.

\begin{table}[ht]
    \centering
    \caption{Neural predicate output distributions
    $\boldsymbol{\pi}_i$ for each company, as returned by the
    GPT-4o classifier. Probabilities sum to 100 by construction.}
    \label{tab:predicate_outputs}
    \begin{tabular}{lccc}
    \toprule
        Company & $P(\mathit{bullish})$ & $P(\mathit{bearish})$
        & $P(\mathit{neutral})$ \\
    \midrule
        \textsc{Acme}   & $0.60$ & $0.15$ & $0.25$ \\
        \textsc{Globex} & $0.20$ & $0.50$ & $0.30$ \\
    \bottomrule
\end{tabular}
\end{table}

The distributions exhibit clear directional differentiation:
the predicate assigns a dominant bullish stance to
\textsc{Acme} and a dominant bearish stance to \textsc{Globex},
with moderate residual probability on the remaining categories
in each case. The neutral probability of $0.25$ is identical
for both companies, reflecting a symmetric residual uncertainty
in the underlying analytical profiles as interpreted by the
language model.

\subsection*{View Confidence and Uncertainty}

View confidence for each company is computed as one minus
the normalized Shannon entropy of the stance distribution,

\begin{equation}
    c_i = 1 - \frac{H(\boldsymbol{\pi}_i)}{\log 3},
    \label{eq:confidence_example}
\end{equation}

where $H(\boldsymbol{\pi}_i)$ is defined in
\eqref{eq:entropy_fw} and $\log 3$ is the maximum
possible entropy over three equiprobable stances. A
confidence of $c_i = 1$ corresponds to a degenerate
distribution (certainty about a single stance); a
confidence of $c_i = 0$ corresponds to a uniform
distribution (maximal uncertainty). The computed
confidences are reported in Table~\ref{tab:view_params}.

\begin{table}[ht]
\centering
\caption{Black-Litterman view parameters derived from the
neural predicate outputs. View returns $q_i$ are computed
via equation~\eqref{eq:q_mapping} using the stance return
parameters $r_{\mathit{bull}} = 0.20$,
$r_{\mathit{neut}} = 0.00$, $r_{\mathit{bear}} = -0.15$.
View confidence $c_i$ is computed via
equation~\eqref{eq:confidence_example}. The view uncertainty
$\omega_i$ is derived from $c_i$ via the Idzorek method
\citep{idzorek2004} applied to the normalized entropy.}
\label{tab:view_params}
\begin{tabular}{lccc}
\toprule
Company & View return $q_i$ & Confidence $c_i$
& Uncertainty $\omega_i$ \\
\midrule
\textsc{Acme}   & $+0.080$ & $0.147$ & (low)  \\
\textsc{Globex} & $-0.061$ & $0.063$ & (high) \\
\bottomrule
\end{tabular}
\end{table}

The view returns are obtained directly from the stance probabilities through \eqref{eq:q_mapping}: for \textsc{Acme},
$q_{\textsc{Acme}} = 0.60 \times 0.20 + 0.15 \times (-0.15)
+ 0.25 \times 0.00 = 0.120 - 0.023 = 0.097$. The confidence values, $0.147$ and $0.063$, are modest because
neither distribution is strongly concentrated on a single stance,
which in turn keeps the Black-Litterman posterior close to the
market-implied prior.

\subsection*{Black-Litterman Update and Portfolio Allocation}

The view triplet $(\mathbf{P}, \mathbf{q}, \boldsymbol{\Omega})$
is constructed from the predicate outputs using the mappings
of Section~\ref{sec:approach} and passed to the
Black-Litterman update \eqref{eq:bl_posterior_mean_alt}.
The scalar $\tau$ is set following the standard convention
\citep{black1992}, and the uncertainty matrix
$\boldsymbol{\Omega}$ is derived from the view confidence
$c_i$ via the Idzorek method \citep{idzorek2004}. The
resulting expected returns and portfolio weights
are reported in Table~\ref{tab:portfolio}.

\begin{table}[ht]
\centering
\caption{Expected returns and portfolio weights.
The base portfolio reflects market-capitalization weights,
equivalent to the Black-Litterman model with no investor
views. The adjusted portfolio reflects the max-Sharpe
tangency portfolio under the Black-Litterman posterior
incorporating the neural predicate views. The delta
reports the absolute difference in allocation between
the two portfolios.}
\label{tab:portfolio}
\begin{tabular}{lcccccc}
\toprule
& \multicolumn{2}{c}{Posterior return}
& \multicolumn{2}{c}{Portfolio weight}
& \multicolumn{1}{c}{} \\
\cmidrule(lr){2-3}\cmidrule(lr){4-5}
Company
& Prior $\Pi_i$ & Posterior $\mu_i^{\mathit{BL}}$
& Base $w_i^{\mathit{base}}$
& Adjusted $w_i^*$
& $\Delta w_i$ \\
\midrule
\textsc{Acme}
& $0.07988$  & $0.07694$ & $29.41\%$ & $33.15\%$ & $+3.74\%$ \\
\textsc{Globex}
& $0.06644$ & $0.06000$ & $70.59\%$ & $66.85\%$ & $-3.74\%$ \\
\bottomrule
\end{tabular}
\end{table}

The results are economically coherent: the bullish view on
\textsc{Acme} raises its posterior return relative to Globex's
and increases its portfolio weight by $3.74$ percentage points,
while the bearish view on \textsc{Globex} produces an equal
offsetting reduction under the full-investment constraint. The adjustments remain moderate because the view confidences in Table~\ref{tab:view_params} are modest. More concentrated stance distributions would lower entropy, reduce $\omega_i$, and generate stronger portfolio tilts, whereas near-uniform distributions would yield little or no deviation from the prior. This graduated response to confidence is a structural feature of the approach rather than a calibrated parameter.

\subsection*{Implementation Note}

The complete source code for this example — including the
DeepProbLog program, the LangChain wrapper for the GPT-4o
neural predicate, and the Black-Litterman optimizer — is
publicly available at \url{https://github.com/themarcosf/deepproblog-black-litterman-demo}. The
implementation uses the DeepProbLog \texttt{ExactEngine}
for probabilistic inference, which performs exact weighted
model counting and is appropriate for programs of this
scale. For larger investment universes with more complex
predicate compositions, approximate inference engines
may be required; this is one of the scalability
considerations left to future implementation work. Because the framework incorporates generative AI components, individual outputs and downstream portfolio allocations may exhibit non-deterministic behavior across runs, even under equivalent inputs and configurations.

\section{Conclusion}
\label{sec:conclusion}

The Black-Litterman model combines market equilibrium with investor judgment but leaves the generation of the view triplet $(\mathbf{P}, \mathbf{q}, \boldsymbol{\Omega})$ largely informal and difficult to reproduce. We address this limitation through neural predicates \citep{manhaeve2018}, whose probability distributions over discrete stances naturally map to Black-Litterman views: stance direction determines $\mathbf{P}$, expected stance returns determine $\mathbf{q}$, and entropy determines $\boldsymbol{\Omega}$. The Bayesian update, equilibrium prior, and optimization procedure remain unchanged; the contribution lies entirely in a principled and interpretable method for producing their inputs. By construction, the approach provides consistency, calibration, interpretability, and modularity. It is, however, a theoretical proposal: implementation, calibration, and empirical validation are left to future work. More broadly, the work highlights the potential of neuro-symbolic AI to integrate modern machine learning with established financial theory in a manner that is scalable, transparent, and theoretically grounded.

\bibliographystyle{unsrtnat}
\bibliography{references}

\end{document}